\documentclass{article}

\usepackage[nonatbib,preprint]{neurips_2021}

\usepackage[utf8]{inputenc} 
\usepackage[T1]{fontenc}    
\usepackage{hyperref}       
\usepackage{url}            
\usepackage{booktabs}       
\usepackage{amsmath,amssymb,amsfonts}       
\usepackage{nicefrac}       
\usepackage{microtype}      
\usepackage{xcolor}         
\usepackage{graphicx}
\usepackage{verbatim}
\usepackage[ruled]{algorithm2e}
\usepackage{algpseudocode}
\usepackage{wrapfig}
\usepackage{subcaption}
\usepackage{soul}

\title{Sensitivity-Aware Finetuning for Accuracy Recovery on Deep Learning Hardware}

\author{%
  Lakshmi Nair\\
  Lightmatter\\
  100 Summer Street\\
  Boston, MA 02110 \\
  \texttt{lakshmi@lightmatter.co} \\
  \And
  Darius Bunandar \\
  Lightmatter\\
  100 Summer Street\\
  Boston, MA 02110 \\
  \texttt{darius@lightmatter.co} \\
}

\begin{document}

\maketitle

\begin{abstract}
Existing methods to recover model accuracy on analog-digital hardware in the presence of quantization and analog noise include noise-injection training. However, it can be slow in practice, incurring high computational costs, even when starting from pretrained models. We introduce the Sensitivity-Aware Finetuning (SAFT) approach that identifies noise sensitive layers in a model, and uses the information to freeze specific layers for noise-injection training. Our results show that SAFT achieves comparable accuracy to noise-injection training and is 2$\times$ to 8$\times$ faster.
\end{abstract}

\section{Introduction}
Recent advances in analog-digital hardware is motivated by improving energy and speed efficiency for deep learning applications. However, such devices are often susceptible to effects of analog noise and reduced precision (quantization) which impacts the final model accuracy. One of the commonly used approaches for tackling this issue includes noise-injection training. Here, the model is subjected to the perturbations caused by quantization and/or analog noise, by injecting some representative noise into the model's layers during training, to recover accuracy \cite{joshi2020accurate, zhou2020noisy, basumallik2022adaptive, baskin2021uniq}. Prior work has shown that loss of precision due to quantization can also be treated as ``noise'' and that models can be made resilient to this loss by retraining with noise injection, where the injected noise is proportional to the precision loss \cite{baskin2021uniq, basumallik2022adaptive}. However, noise-injection training can incur significant training time, even when starting from pretrained models. The speed of training can be potentially improved by training only a subset of the layers that are highly sensitive to noise, while freezing (i.e., disable weight updates) the rest. As shown in Figure \ref{fig:weight_figure} for ResNet50, some weights change substantially during noise-injection training, while others hardly change and could potentially be frozen (conceptually similar to transfer learning \cite{zhuang2020comprehensive}). While prior work has looked at similar methods for speeding up training, they either focus on BERT-like models \cite{vucetic2022efficient, zaken2021bitfit}, or rely on domain knowledge to identify noise sensitive layers \cite{piao2022sensimix}.


Motivated by these observations, we seek to answer the question: ``\textit{Starting from a pretrained model, can we identify which layers are the most sensitive to noise, and retrain just those?}''. Prior work in quantization has used KL-divergence to identify layers that are sensitive to quantization \cite{tensorrt,kummer2021adaptive,gholami2021survey}. We present an alternate metric for measuring layer sensitivity to noise, by computing the \textit{standard deviation} of the output differences between the noisy/quantized model and the noise-free/unquantized model at each layer. We then introduce the Sensitivity-Aware Finetuning (SAFT) approach based on noise-injection training, that selects specific layers for training based on the layer sensitivity analysis.

\section{Sensitivity-Aware Finetuning}



\begin{figure}
\centering
\begin{subfigure}{0.32\textwidth}
  \centering
  \includegraphics[width=\linewidth]{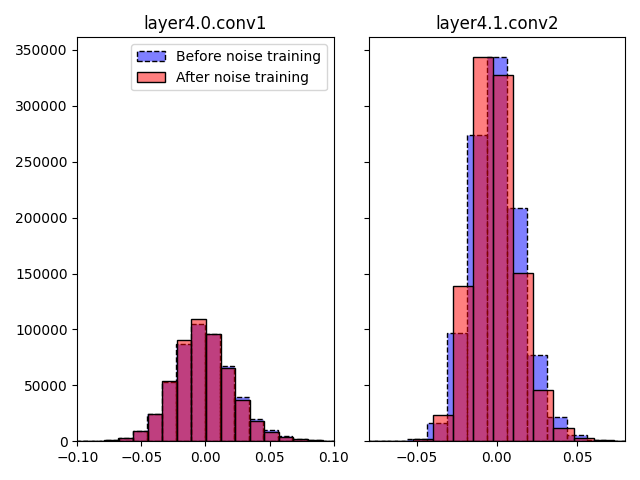}
  \caption{Weights of ResNet50, before and after noise-injection training from a pretrained model. For some layers, the weights do not change much post-training.}
  \label{fig:weight_figure}
\end{subfigure}%
\hspace{4mm}
\begin{subfigure}{0.64\textwidth}
  \centering
  \includegraphics[width=\linewidth]{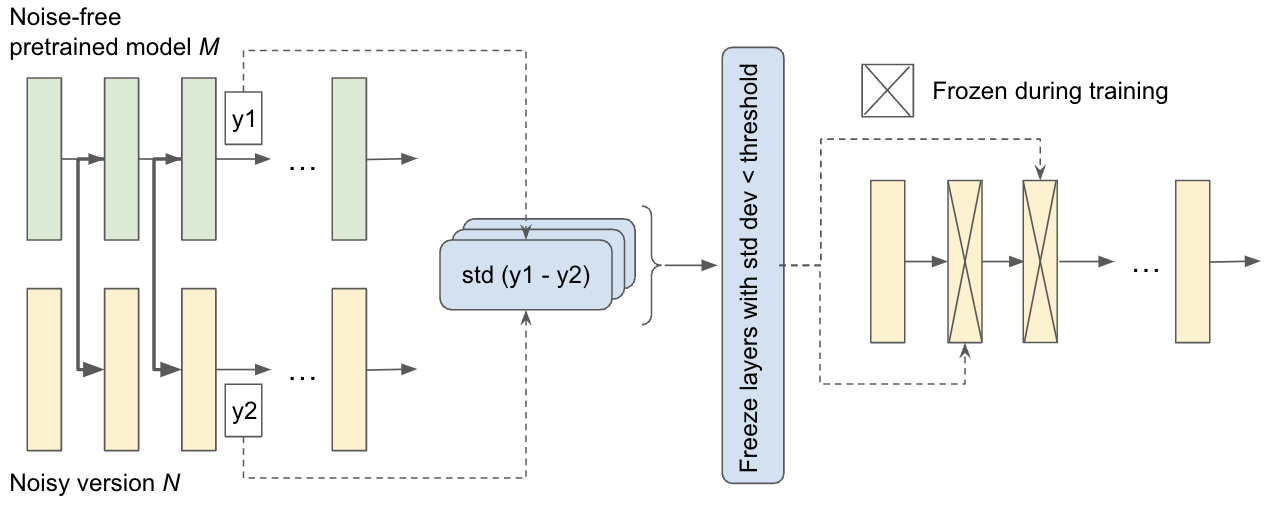}
  \caption{SAFT begins by passing the inputs of each layer in the noise-free model through the corresponding layers of the noisy model. The outputs for both models at each layer is used to compute a standard deviation over their differences. Only layers with standard deviations above a certain threshold are trained using noise-injection training while the remaining are frozen.}
  \label{fig:SAFT-flow}
\end{subfigure}
\caption{Sensitivity-Aware Finetuning (SAFT): motivation (left) and description (right)}
\label{fig:saft-des-mot}
\end{figure}


The motivation behind SAFT is the observation that after noise-injection training, the parameters of a model change significantly only for specific layers, e.g., Figure \ref{fig:weight_figure} shows this for ResNet50. Keeping such layers noise-free in the model results in larger accuracy improvements. This possibly indicates that the layers whose parameters change the most have higher noise sensitivity. Hence, we seek to retrain only the most noise sensitive layers to validate this hypothesis.

Our approach takes a pretrained model $M$, its noisy version $N$, and a sample batch of inputs $X$ from the training data. Note that $N$ refers to the model used during standard noise-injection training \cite{joshi2020accurate}, where we inject noise into the weights during the forward pass to perturb the outputs. The input data is first passed through $M$, and the inputs and outputs at every layer are stored. Then, the inputs at every layer of the \textit{original model} $M$ are passed through the \textit{corresponding layers of} $N$ (See Appendix Algorithm 1). The layer outputs for $N$ are also saved, and the standard deviation\footnote{Noise mean is typically zero based on hardware models \cite{garg2021dynamic, joshi2020accurate, basumallik2022adaptive}} of the differences between the outputs of $M$ and $N$ are computed per-layer. The process flow is shown in \ref{fig:SAFT-flow}.

Once the standard deviations are computed, SAFT involves: a) Identifying the top $k$ layers with the highest standard deviations; b) Selectively training only the top $k$ layers of $N$ while freezing the parameters of the remaining layers. We note that $k$ is an additional hyperparameter. The value of $k$ can be determined based on visualizing the standard deviation values in a plot, or it can be treated as the other hyperparameters and set using tools such as Tune \cite{liaw2018tune}. Another consideration here is the batch size used for computing the statistics. The batch size should be sufficiently large to obtain a reasonable estimate of the noise sensitivity\footnote{We find that using the batch sizes typically used for training, works well in most cases}. When large batches cannot be processed, data samples can be processed individually and stacked. The statistics can then be accumulated over the stack. For training, we use the procedure in \cite{joshi2020accurate}, and apply the backward gradient updates to noise-free weights.

\textbf{Computational complexity of standard deviation based layer sensitivity analysis:} Existing layer sensitivity approaches start with the un-quantized model and proceed by quantizing a single layer at a time, evaluating the model accuracy in each case \cite{wu2020integer, openvino}. The layers are then sorted in decreasing order of sensitivity and the most sensitive layers are skipped during quantization \cite{wu2020integer}. Existing software packages such as OpenVino have introduced the Accuracy Aware Algorithm, a slightly modified version of the approach starting with the quantized version of the model \cite{openvino}. However, these approaches are brute-force methods that can take $O(Nt)$ time for a model with $N$ layers and evaluation time $t$ for the sample of data. In contrast, the standard deviation based sensitivity analysis only requires a sample of data to be passed \textit{once} through the model for a reduced complexity of $O(t)$. For models with large $N$ this can lead to significant speed improvements in layer sensitivity analysis.




\section{Experiments}
We evaluate SAFT on eight different models. In all cases, similar to prior work in \cite{wu2020integer} we only apply noise to the matrix-multiplication layers (such as Convolutions, Linear etc.), and leave other layers such as batchnorm or activation layers as noise-free. We also evaluate the use of KL-divergence as an alternate metric to standard deviation in our experiments. Similar to prior work, we evaluate SAFT with simulated hardware noise using both multiplicative and additive noise, wherein noise is injected into the weights \cite{joshi2020accurate, zhou2020noisy, tsai2020robust}. We sample the noise $N$ from both a Gaussian distribution with zero mean as in prior work \cite{zhou2020noisy}, $N \sim \mathcal{N}(0, \sigma)$, and from a Uniform distribution $N \sim U[-r_1, r_1]$. Our baseline noise-injection is implemented similar to the approach in \cite{joshi2020accurate}. The parameters of the noise distributions for the different models are shown in Appendix Table \ref{tab:noise_params}. The specific noise parameters were chosen so as to result in a drop in the performance of all the models, which can then be recovered through training. Note that we set a fixed seed for all our training runs to ensure fair comparison.

For SAFT, we compute the standard deviation values on a single batch of training data. We freeze $total - k$ layers in a model during training, retraining only $k$. We determined $k$ empirically by visualizing the standard deviation plots and checking the number of layers that have a relatively high noise standard deviation. We seek to standardize this procedure in our future work. Table \ref{tab:speedup_perf} in the Appendix shows the batch size, and $k$ values ($\# \text{Frozen} = \# \text{Total} - k$) used in our experiments. Some models require more layers to be trained than others owing to higher noise in more layers.

Our experiments evaluate: \textit{Given the exact same training parameters, does SAFT perform similar to baseline noise-injection training?} Note that our research question compares our approach to noise-injection training, rather than to obtain a predefined target performance, which noise-injection training has already been shown to achieve with sufficient epochs \cite{joshi2020accurate, baskin2021uniq, wu2020integer}. In our training experiments, we only train for a few epochs (1-5 epochs) to see if the performances of the two approaches match, whereas achieving close to the baseline noise-free FP32 performance takes many more epochs \cite{wu2020integer}.

\section{Results}

\setlength{\tabcolsep}{3.8pt}
\begin{table}[]
\caption{Results comparing SAFT with baseline noise-injection training with Gaussian noise shows similar performances. Here ``Untrained'' denotes performance before training, when noise is injected. Note that SAFT achieves accuracy close to noise-injection training while being 2$\times$ to 8$\times$ faster.}
\label{tab:gauss_perf}
\begin{tabular}{lc|ccc|ccc|c}
& \multicolumn{1}{l|}{} & \multicolumn{3}{c|}{\textbf{Multiplicative Gaussian}} & \multicolumn{3}{c|}{\textbf{Additive Gaussian}} & \textbf{SAFT}\\ \multicolumn{1}{l|}{\textbf{}} & \multicolumn{1}{l|}{\textbf{FP32}} & \multicolumn{1}{l}{\textbf{Untrained}} & \multicolumn{1}{l}{\textbf{Noise-inj}} & \multicolumn{1}{l|}{\textbf{SAFT}} & \multicolumn{1}{l}{\textbf{Untrained}} & \multicolumn{1}{l}{\textbf{Noise-inj}} & \multicolumn{1}{l|}{\textbf{SAFT}} &
\multicolumn{1}{l}{\textbf{Speed $\uparrow$}}\\
\multicolumn{1}{l|}{\textbf{ResNet18}}  & 69.8  & 68.7 & 69.1  & 69.0  & 66.0 & 67.4 & 67.8 & 2$\times$ \\
\multicolumn{1}{l|}{\textbf{ResNet34}}  & 73.3  & 72.1 & 72.9  & 72.9  & 69.0 & 70.0 & 70.0 & 4$\times$ \\
\multicolumn{1}{l|}{\textbf{ResNet50}} & 76.1 & 74.9 & 75.8 & 75.6 & 70.7 & 73.1 & 73.1 & 8$\times$ \\
\multicolumn{1}{l|}{\textbf{ResNeXt50}} & 77.6 & 72.2 & 74.4 & 74.0  & 71.1 & 73.9 & 74.2 & 8$\times$ \\
\multicolumn{1}{l|}{\textbf{MobileNet v3}} & 74.0 & 70.8 & 71.7 & 71.6 & 70.9 & 72.7 & 72.6 & 5$\times$ \\
\multicolumn{1}{l|}{\textbf{Faster RCNN}} & 59.0 & 56.5 & 58.9 & 58.7 & 52.2 & 54.4 & 54.8 & 3$\times$ \\
\multicolumn{1}{l|}{\textbf{Mask RCNN}} & 56.0 & 52.0 & 55.3 & 55.6 & 48.5 & 53.6 & 54.9 & 3$\times$ \\
\multicolumn{1}{l|}{\textbf{Bert Base}} & 74.7 & 73.1 & 74.4 & 74.6 & 72.4 & 74.4 & 74.2 & 2$\times$     
\end{tabular}
\end{table}

\begin{table}[]
\caption{Results comparing SAFT with baseline noise-injection training for Uniform noise shows similar performances. Here ``Untrained'' denotes performance before training, when noise is injected.}
\label{tab:uniform_perf}
\centering
\begin{tabular}{lc|ccc|ccc|c}
& \multicolumn{1}{l|}{} & \multicolumn{3}{c|}{\textbf{Multiplicative Uniform}} & \multicolumn{3}{c|}{\textbf{Additive Uniform}} & \textbf{SAFT} \\ \multicolumn{1}{l|}{\textbf{}} & \multicolumn{1}{l|}{\textbf{FP32}} & \multicolumn{1}{l}{\textbf{Untrained}} & \multicolumn{1}{l}{\textbf{Noise-inj}} & \multicolumn{1}{l|}{\textbf{SAFT}} & \multicolumn{1}{l}{\textbf{Untrained}} & \multicolumn{1}{l}{\textbf{Noise-inj}} & \multicolumn{1}{l|}{\textbf{SAFT}} & \textbf{Speed} $\uparrow$\\
\multicolumn{1}{l|}{\textbf{ResNet18}}  & 69.8  & 68.2 & 69.3  & 69.0  & 64.8 & 66.9 & 66.2 & 2$\times$ \\
\multicolumn{1}{l|}{\textbf{ResNet34}}  & 73.3  & 71.8 & 72.7  & 72.6  & 67.1 & 68.5 & 68.0 & 4$\times$ \\
\multicolumn{1}{l|}{\textbf{ResNet50}} & 76.1 & 74.5 & 75.1 & 75.3  & 67.9 & 70.5 & 70.1 & 8$\times$ \\
\multicolumn{1}{l|}{\textbf{ResNeXt50}} & 77.6 & 71.3 & 73.0 & 73.1  & 73.0 & 75.7 & 75.3 & 8$\times$ \\
\multicolumn{1}{l|}{\textbf{MobileNet v3}} & 74.0 & 71.4 & 72.1 & 72.2 & 72.9 & 73.9 & 73.8 & 5$\times$ \\
\multicolumn{1}{l|}{\textbf{Faster RCNN}} & 59.0 & 57.1 & 58.2 & 58.2 & 56.2 & 57.0 & 57.2 & 3$\times$ \\
\multicolumn{1}{l|}{\textbf{Mask RCNN}} & 56.0 & 53.4 & 54.5 & 54.8 & 49.5 & 51.5 & 51.6 & 3$\times$ \\
\multicolumn{1}{l|}{\textbf{Bert Base}} & 74.7 & 68.9 & 72.0 & 72.4 & 62.2 & 72.1 & 72.7 & 2$\times$     
\end{tabular}
\end{table}

\begin{figure}[t]
\centering \includegraphics[width=0.74\textwidth]{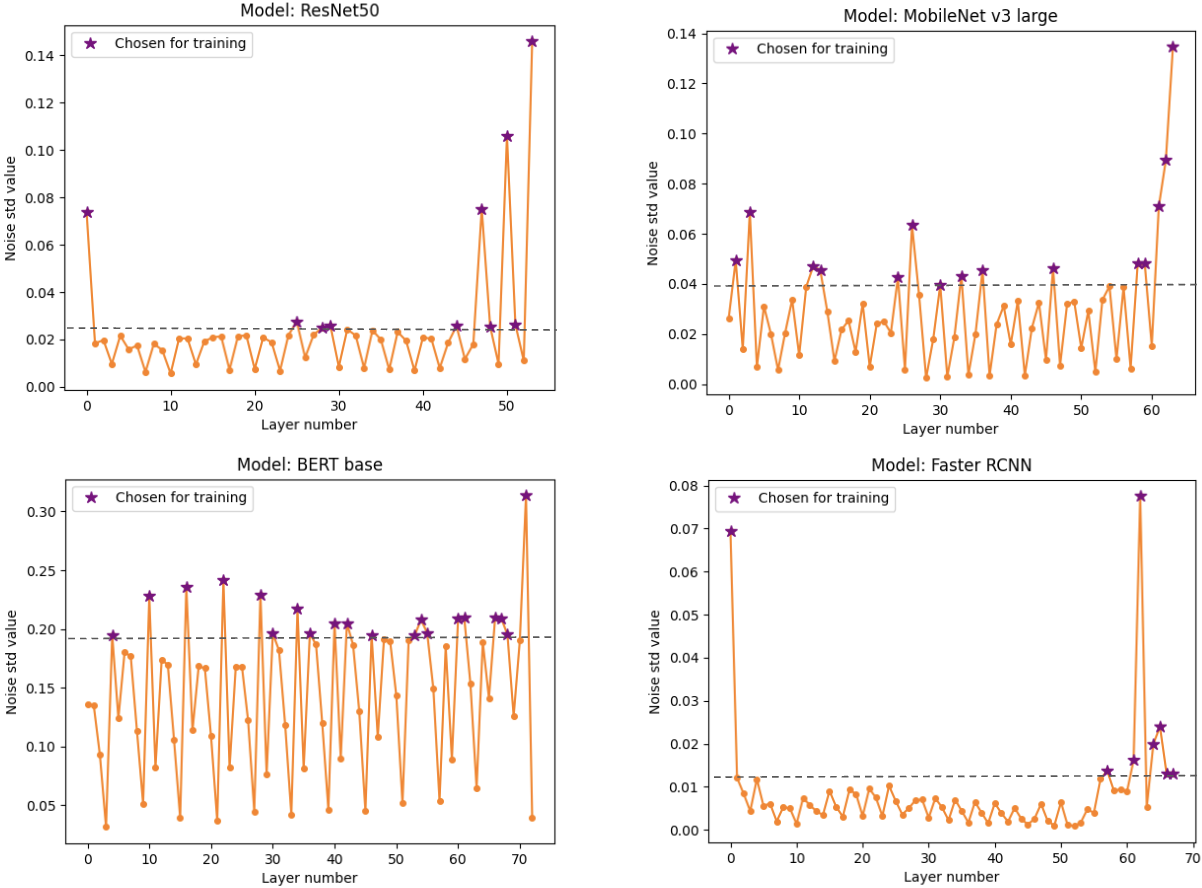}
\caption{Plots of the per-layer standard deviations for four models: ResNet50, MobileNet v3 large, Bert base, and Faster RCNN. The purple stars $\star$ denote the layers that were selected for training while the rest were frozen. First and last layers of vision models have high standard deviations.}
\label{std_figure}
\end{figure}

\begin{table}[]
\caption{Results for finetuning with Gaussian injected noise using KL-divergence to freeze layers, compared with using standard deviation to select the layers to freeze. Standard deviation based layer freezing (shown in bold) outperforms layer selection using KL-divergence.}
\label{tab:KL_training}
\centering
\begin{tabular}{l|c|c|c}
\textbf{} & \textbf{Baseline FP32} & \textbf{Mult / Add (Std)} & \textbf{Mult / Add (KL-d)} \\ 
\multicolumn{1}{l|}{\textbf{ResNet34}} & 73.3 & \textbf{72.9 / 70.0} & 71.9 / 68.2 \\
\multicolumn{1}{l|}{\textbf{ResNet50}} & 76.1 & \textbf{75.6 / 73.1} & 74.2 / 69.4 \\
\multicolumn{1}{l|}{\textbf{Bert base}} & 74.7 & \textbf{74.6 / 74.2} & 74.2 / 74.1 \\
\end{tabular}
\end{table}

Noise standard deviation plots for four models are shown in Figure \ref{std_figure}. Stars $\star$ indicate the layers that are trained, while the remaining are frozen. Similar to prior findings, the first and last set of layers in vision models exhibit high sensitivity \cite{mckinstry2018discovering,rastegari2016xnor,joshi2020accurate}. We also see a ``sawtooth'' pattern in the vision models like ResNet, corresponding to the repeating blocks in the network, consistent with observations in prior work \cite{garg2021dynamic}. For MobileNet v3, quite a few of the convolution layers have a higher noise standard deviation compared to ResNet50. For Faster RCNN, we see that several layers in the ``head'' of the model, responsible for predicting the bounding box locations, are particularly sensitive. For Bert base, the sensitivity is quite spread out across the model with several layers exhibiting high noise sensitivity. Specifically, 10 of the trained 20 layers are self-attention layers, with the remaining 10 being intermediate and output dense layers. For models like Bert, identifying the most sensitive layers can be tricky and a larger proportion of layers have to be trained compared to other models. 

The corresponding training speed improvements are shown in Tables \ref{tab:gauss_perf} and \ref{tab:uniform_perf}, where up to $8\times$ speed improvements in training can be observed. In the case of a few models such as ResNet18 and RCNN, speed up of about $2\times$ and $3\times$ is observed. The actual amount of speedup depends on the processing time of each layer, which in turn depends on the size of the layer (i.e., \# of parameters). Hence, a direct correlation between size of the model (i.e., \# of layers) and the speedup is difficult to establish.

The final performance of SAFT in terms of the model metrics is shown in Tables \ref{tab:gauss_perf} and \ref{tab:uniform_perf} for Gaussian and Uniform noise respectively. We see that SAFT (with $k$ frozen layers) closely matches\footnote{Our results were confirmed with the Wilcoxon Signed Rank test ($\alpha = 0.05$)} the performance of the full noise-injection training approach for all noise models, leading to improvements in terms of the metrics. An interesting finding here is that specific layers that do not form a continuous sequence, can be independently trained. Typically in transfer learning a \textit{continuous} sequence of the last few layers, such as the last few convolutional and fully-connected layers, are often retrained \cite{zhuang2020comprehensive}.

Lastly, we train a few models with Gaussian noise injection, by using KL-divergence for selecting the layers to freeze as opposed to using standard deviation (See Table \ref{tab:KL_training}). The results clearly show that freezing layers based on their standard deviations outperforms KL-divergence based layer selection. Furthermore, computation of KL-divergence requires converting the activations into probability distributions, which can be avoided in the case of the standard deviation based method. Interestingly, using KL-divergence did not improve performance on vision models, although it did perform well on Bert base. It is possible that since most layers in Bert base have high noise sensitivity (See Figure \ref{std_figure}), KL-divergence chose and trained some of the noisiest layers, whereas the noisiest layers are much more specific and localized in the case of the vision models. Especially in such cases, KL-divergence could not perform as well as standard deviation in identifying these more specific layers.

\section{Conclusions and Future Work}
We introduced Sensitivity-Aware Finetuning (SAFT) for fast finetuning of pretrained models to deal with noise. SAFT computes layer sensitivity using standard deviations to freeze some layers. SAFT performs comparably to noise-injection training in terms of accuracy, while being faster at training. In the future, we will investigate additional metrics for SAFT, including combinations of metrics like standard deviation and KL-divergence. We will investigate techniques for easily identifying the $k$ hyperparameter used in SAFT. We believe the layer sensitivity analysis can also be used for performing Partial Quantization and Quantization-Aware Training \cite{wu2020integer} in future work.

\bibliography{main}
\bibliographystyle{IEEEtran}

\appendix

\section{Appendix}

\begin{table}[h]
\caption{\# of total layers vs. frozen layers, and the speed increase for SAFT over baseline noise-injection training (*Lower $k$ values could potentially be obtained through a more rigorous search).}
\label{tab:speedup_perf}
\centering
\begin{tabular}{l|c|c|c|c}
\textbf{} & \textbf{Batch size} & \textbf{k} & \textbf{\# Total} & \textbf{\# Frozen} \\ 
\multicolumn{1}{l|}{ResNet18} & 256 & 10 & 21 & 11 \\
\multicolumn{1}{l|}{ResNet34} & 256 & 10 & 37 & 27 \\
\multicolumn{1}{l|}{ResNet50} & 256 & 10 & 54 & 44 \\
\multicolumn{1}{l|}{ResNeXt50} & 256 & 10 & 54 & 44 \\
\multicolumn{1}{l|}{MobileNet v3} & 256 & 15 & 64 & 49 \\
\multicolumn{1}{l|}{Faster RCNN} & 20 & 8 & 68 & 60 \\
\multicolumn{1}{l|}{Mask RCNN} & 20 & 8 & 74 & 66 \\
\multicolumn{1}{l|}{Bert base} & 80 & 20 & 73 & 53 \\
\end{tabular}
\end{table}

\begin{table}[h]
\centering
\caption{Noise distribution parameters for the different models. The specific noise parameters were chosen so as to result in a drop in the performance of all the models, which can then be recovered through training.}
\label{tab:noise_params}
\begin{tabular}{l|c|c|c|c|c} & \multicolumn{2}{c|}{\textbf{Gaussian Noise $\sigma$}} & \multicolumn{2}{c}{\textbf{Uniform noise $r_1$}} \\
\textbf{} & \textbf{Mult} & \textbf{Add} & \textbf{Mult} & \textbf{Add} & \textbf{Epochs (training)}\\
\textbf{ResNet18} & 0.05 & 0.005 & 0.1 & 0.01 & 3\\
\textbf{ResNet34} & 0.05 & 0.005 & 0.3 & 0.01 & 3\\
\textbf{ResNet50} & 0.05 & 0.005 & 0.3 & 0.01 & 3\\
\textbf{ResNeXt50} & 0.08 & 0.005 & 0.15 & 0.008 & 3\\
\textbf{MobileNet v3 large} & 0.02 & 0.005 & 0.03 & 0.008 & 5\\
\textbf{Mask RCNN} & 0.01 & 0.005 & 0.01 & 0.008 & 1\\
\textbf{Faster RCNN} & 0.05 & 0.005 & 0.05 & 0.005 & 1\\
\textbf{Bert base} & 0.02 & 0.002 & 0.3 & 0.01 & 1        
\end{tabular}
\end{table}

\begin{algorithm}
\label{alg:layer_ssty}
\caption{Computing layer sensitivity}
\SetKwInput{KwInput}{Input}
\SetKwInput{KwOutput}{Output}
\SetKwInput{KwData}{Data}
\DontPrintSemicolon
  
\KwInput{Original model $M$, noisy model $N$}
\KwOutput{Layer output standard deviations $S$}
\KwData{Sample batch of inputs $X$}

  \SetKwFunction{FStat}{compute\_stats}
  \SetKwFunction{FMain}{Main}
  \SetKwFunction{FStore}{save\_data}
 
  \SetKwProg{Fn}{Function}{:}{}
  \Fn{\FStore{$M$, $X$}}{
    \tcp{Save input-output of each layer}
        $I_{model}$ = \{\}\;
        $O_{model}$ = \{\}\;
        \For{i, l in enumerate(M.layers)}{
            \eIf{i = 0}{
                $Y$ = $l(X)$\;
                $I_{model}[l]$ = $X$\;
                $O_{model}[l]$ = $Y$\;
            }{
            $I_{model}[l]$ = $Y$\;
            $Y$ = $l(Y)$\;
            $O_{model}[l]$ = $Y$\;
            }
        }
        \KwRet $I_{model}$, $O_{model}$\;
  }
  \;

  \SetKwProg{Fn}{Function}{:}{}
  \Fn{\FStat{$M$, $N$, $X$}}{
        $I_{clean}, O_{clean}$ = \FStore{$M$, $X$}\;
        $S$ = \{\}\;
        \For{l in N.layers}{
        \tcp{Pass inputs of M through N}
            $Y_{noisy} = l(I_{clean}[l])$\;
            $\hat{Y} = Y_{noisy} - O_{clean}[l]$\;
            $S[l]$ = $std(\hat{Y})$\;
        }
        \tcp{Sort in decreasing std values}
        \KwRet $sort(S, \downarrow)$\;
  }
  \;
\end{algorithm}

\end{document}